\documentclass[10pt,twocolumn,letterpaper]{article}

\usepackage{wacv}              

\usepackage{graphicx}
\usepackage{amsmath}
\usepackage{amssymb}
\usepackage{booktabs}
\usepackage{graphicx}
\usepackage{float}
\usepackage{listings}
\usepackage{multirow}
\usepackage[margin=3cm]{geometry}
\usepackage{algorithm2e}
\usepackage{amsfonts,mathtools}
\usepackage[table,xcdraw]{xcolor}

\usepackage[many]{tcolorbox}        

\definecolor{main}{HTML}{5989cf}    

\tcbset{
    sharp corners,                  
    colback = white,                
    before skip = 0.2cm,            
    after skip = 0.5cm              
}

\newtcolorbox{boxK}{
    sharpish corners,               
    boxrule = 0pt,                  
    toprule = 4.5pt,                
    enhanced,                       
    fuzzy shadow = {0pt}{-2pt}{-0.5pt}{0.5pt}{black!35} 
}

\usepackage[pagebackref,breaklinks,colorlinks]{hyperref}

\usepackage[capitalize]{cleveref}
\crefname{section}{Sec.}{Secs.}
\Crefname{section}{Section}{Sections}
\Crefname{table}{Table}{Tables}
\crefname{table}{Tab.}{Tabs.}

\def\wacvPaperID{*****} 
\def\confName{WACV}
\def\confYear{2025}

\begin{document}

\title{Lights, Camera, Matching: \\ The Role of Image Illumination in Fair Face Recognition}

\author{
    Gabriella Pangelinan$^{1}$, Grace Bezold$^{2}$, Haiyu Wu$^{2}$, Michael C. King$^{1}$, Kevin W. Bowyer$^{2}$\\
    \begin{minipage}[t]{0.45\textwidth}
        \centering
        $^{1}$Florida Institute of Technology\\
        Melbourne, FL, USA\\
        {\tt\small gpangelinan@my.fit.edu \\ michaelking@fit.edu}
    \end{minipage}%
    \hfill
    \begin{minipage}[t]{0.45\textwidth}
        \centering
        $^{2}$University of Notre Dame\\
        Notre Dame, IN\\
        {\tt\small \{gbezold,hwu6,kwb\}@nd.edu }
    \end{minipage}
}

\maketitle

\begin{abstract}
    Facial brightness is a key image quality factor impacting face recognition accuracy differentials across demographic groups. In this work, we aim to decrease the accuracy gap between the similarity score distributions for Caucasian and African American female mated image pairs, as measured by d' between distributions. To balance brightness across demographic groups, we conduct three experiments, interpreting brightness in the face skin region either as median pixel value or as the distribution of pixel values. Balancing based on median brightness alone yields up to a 46.8\% decrease in d', while balancing based on brightness distribution yields up to a 57.6\% decrease. In all three cases, the similarity scores of the individual distributions improve, with mean scores maximally improving 5.9\% for Caucasian females and 3.7\% for African American females.
   
\end{abstract}

\section{Introduction}\label{introduction}

Various research groups have shown
that face recognition accuracy varies across demographic groups \cite{bias-survey, klare2012face, abdurrahim2018review, frvt_dem_effects}. 
Comparing the distributions of similarity scores for either mated image pairs or non-mated image pairs, accuracy tends to be higher for males than females \cite{vitor-gender-differences, frvt-gender}. Comparing accuracy for Caucasians versus African Americans, the distribution of similarity scores for non-mated image pairs tends to be better for Caucasians, and the distribution of similarity scores for mated pairs tends to be better for African Americans \cite{vangara2019characterizing,cavazos2020accuracy}. 


Diverse methods for mitigating the accuracy gap between Caucasians and African Americans have been proposed. Some researchers have focused on modifying model configuration, e.g. by normalizing features \cite{terhorst2020post}, introducing novel loss functions \cite{terhorst2020comparison}, or implementing demographic-adaptive architecture \cite{gong2021mitigating}. Others have addressed the issue of imbalanced training and testing data, e.g. with respect to representation of demographic groups \cite{wu2023should,albiero2020does} or appearance-based factors like facial hair \cite{ozturk2024can,wu2024facial,bhatta2023gender}. 
(Interestingly, demographic balance of number of identities / images in the training data does not appear to be an effective lever for balanced accuracy on test data \cite{albiero2020does}.)
Still others have targeted image-based quality factors (e.g. illumination, resolution, and blur) related to the image acquisition environment. 

In this work, we focus on illumination of the face, in terms of the grayscale brightness of the face region of an image. 
Previous works exploring brightness as a quality measure have generally considered 
an overall brightness value for each image independently.
Here, we 
consider how brightness impacts a \textit{pair} of images involved in a mated image pair; that is, a pair of images of the same person. 

We perform three experiments using a dataset of Caucasian female (CF) and African American female (AF) images, with the goal of balancing illumination of the face region across race to see how this impacts the accuracy gap. The images in this dataset are ``mugshot style'' images, captured 
with controlled illumination and plain background, rather than ``in-the-wild''.
In the first experiment,
we treat face brightness as a discrete value for an image, and consider how the difference in brightness between two images in a mated pair impacts the similarity score used for face recognition. 
%
The next two experiments explore the distributions of brightness values associated with images in a pair. In Sec. \ref{experiment2} we characterize the modality (uni-, bi-, or multimodal) of each distribution. In Sec. \ref{experiment3} we measure the similarity, using intersection-over-union, of  distributions of brightness values across face regions.

Each experiment explores a potential brightness-based ``balancing factor'' - a 
way to
create sets of CF and AF mated pairs that are balanced on face brightness.
We then ask how successful each approach to brightness balancing is at reducing the difference in similarity scores between AF and CF, in comparison to the original unbalanced datasets.

\section{Related Work}\label{related}

In previous works analyzing brightness as a quality factor, the metrics for measuring brightness differ, and their conclusions are not necessarily consistent: e.g. high-brightness face regions degrade performance \cite{terhorst-brightness} \textit{or} higher brightness can increase recognition rate \cite{abaza2014design}.
Additionally, these works generally focus on the brightness of a single image \textit{and} do not explore the effects across demographic groups. Since we are concerned with how brightness impacts accuracy,
\textit{specifically} across demographic groups, we find one relevant (though not brightness-specific) result in a 2022 FRVT report: ``False negatives are in large part due to one or both photographs being of poor quality, something that can be coupled with demographics'' \cite{frvt2022}.

We previously reported \cite{wu2023face} the first results, to our knowledge, that explored how the level of brightness in a \textit{pair} of images, rather than a single image, impacts the corresponding face recognition accuracy for non-mated image pairs (i.e. having different identities) across demographics. We proposed a metric for measuring the brightness of the face skin region as mean pixel value, and based on this value assigned each image an exposure level ranging from strongly under-exposed to strongly over-exposed. With this classification, non-mated image pairs in which the images are either both under-exposed or both over-exposed were found to have a higher False Match Rate (FMR). 
Alternately, pairs in which the exposure level between the two images varies greatly have a lower FMR. For mated pairs, when the exposure level between the two images varies greatly \textit{or} both images are under-/over-exposed, the False Non-Match Rate (FNMR) increases. 

In this work, we focus on mated (``genuine'') score distributions.
(We consider score distributions for female demographics in order to avoid complications related to facial hair.)
We 
explore brightness difference between two images in a pair as a factor affecting similarity scores.
Additionally, we attempt to understand brightness through a more holistic lens, considering how the \textit{shapes} of brightness value distributions (i.e. their modalities) impact similarity scores. 

\section{Overview of Experiments}\label{overview}



Our goal is to explore how balancing the illumination (brightness) of the face region across race impacts the observed difference in similarity scores for mated image pairs.
The baseline result is the observed d’ distance between the distributions of mated scores for AF and CF.
A d’ of zero would represent equal accuracy for AF and CF.
Different approaches to balancing brightness across AF and CF mated pairs are evaluated on the percent by which they shift the baseline d’ (``d' shift''). A negative d' shift indicates that the balanced distributions are closer together than the baseline distributions.

Each experiment uses a different approach to balancing AF and CF image pairs on face brightness.
Balanced subsets of N thousand each of AF and CF image pairs are created by selecting from the original pool of image pairs.
In the two approaches where the pairs can be ordered from well-balanced to less well-balanced, the smaller subsets are increasingly well-balanced.
In the approach where there is no inherent ordering, we shuffle all pairs before taking each set of N thousand, and give average values across 10 trials.
We report the following metrics for each set of N thousand pairs:
\begin{itemize}
    \item The \% change in the mean similarity score of the balanced subset (indicated by $\bar{x}_{b}$) vs. that of the baseline (unbalanced) distribution ($\bar{x}_{u}$).
    \item The \% change from the baseline d' to the d' of the balanced subsets (``balanced d''').
\end{itemize}


\subsection{Dataset and Matcher}\label{dataset}
The MORPH dataset \cite{ricanek2006morph} consists of mugshot images which are (1) taken with nominally controlled lighting, pose, and expression and (2) annotated with race / gender labels. We use the version curated in \cite{vitor-gender-differences}. It contains 10,941 images of 2,798 Caucasian females (``CF'') corresponding to 33,470 mated pairs and 24,857 images of 5,929 African American females (``AF'') corresponding to 82,572 mated pairs.  

Our matcher is a combined margin model based on ArcFace loss~\cite{Deng_CVPR_2019} trained on Glint360k \cite{glint360k} with weights at \cite{Insightface}. The matcher's input is an aligned face image resized to 112x112, and the output is a 512-d feature vector matched using cosine similarity. For each image pair, the matcher provides a cosine similarity score; values closer to +1 indicate greater similarity.

\subsection{Baseline Results}\label{overview-baseline-res}
\begin{figure}[ht!]
  \centering
  \includegraphics[width=\columnwidth]{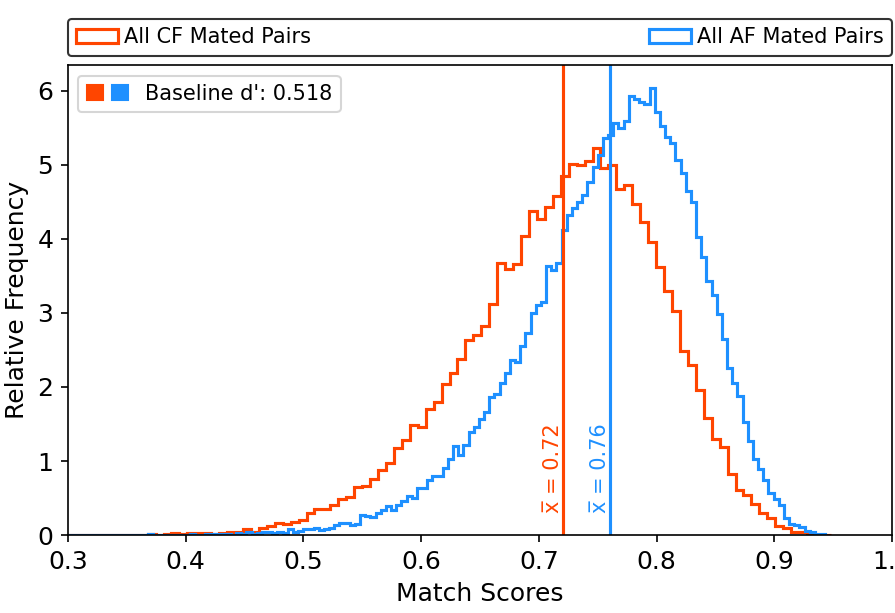}
  \caption{Baseline distributions and $\bar{x}_u$ values for CF / AF. The distribution of CF mated similarity scores is shifted toward lower similarity, relative to the AF distribution.}
  \label{fig:baseline-cf-af}
\end{figure}

The baseline score distributions for CF / AF mated pairs are shown in Fig. \ref{fig:baseline-cf-af}. 
Several results will accomplish our primary goal of selecting subsets of the CF / AF distributions that have a smaller d' than the baseline distributions. In our case, the most preferable option is that \textit{both} the CF and AF distributions shift to higher similarity, with the CF shift being larger. This would mean narrowing the accuracy gap between AF and CF while improving the accuracy of each.
Thus, our secondary goal is ensuring that each subset distribution either shifts to higher similarity scores, or at least does not shift to lower similarity.

Note that we will refer to the mean score of the baseline distributions as $\bar{x}_u$, indicating the ``unbalanced mean'', and the mean score of the balanced distributions as $\bar{x}_b$. The mean value of a balancing factor may also be represented $\bar{x}_b$ - this distinction is clarified as necessary.

\subsection{Data Pre-Processing}\label{overview-preprocessing}
\begin{figure}[!ht]
  \centering
  \includegraphics[width=\columnwidth]{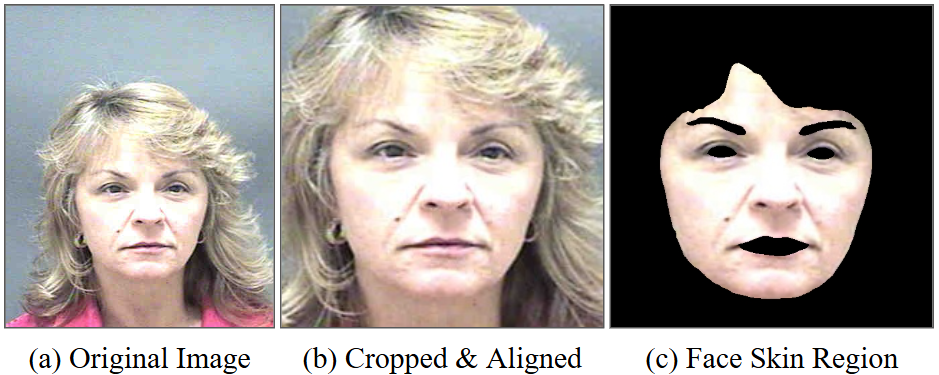}
  \caption{Data pre-processing pipeline.}
  \label{fig:preprocessing-step-1}
\end{figure}

MORPH images are originally 400x480 (Fig. \ref{fig:preprocessing-step-1}a). We use RetinaFace \cite{retinaface} to crop / align the images to 224x224 (Fig. \ref{fig:preprocessing-step-1}b). Then, we use BiSeNet \cite{bisenet} to extract the array of pixels representing the ``face skin region'' (Fig. \ref{fig:preprocessing-step-1}c). The pixel array is converted to grayscale so that each pixel is a single value and represents brightness. 

\section{Brightness Value Difference}\label{experiment1}

\subsection{Pre-Processing}

\begin{figure}[!ht]
  \centering
  \includegraphics[width=\columnwidth]{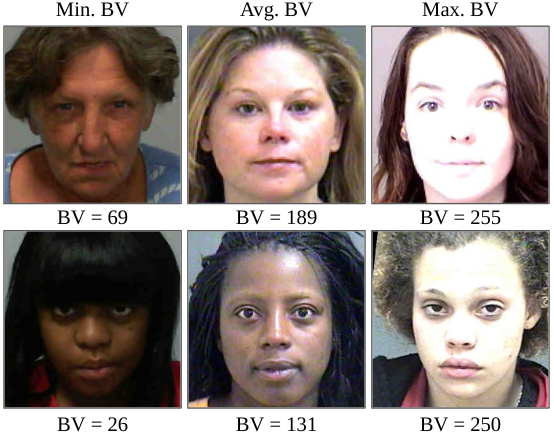}
  \caption{Ex. images with min., avg., and max. BVs.}
  \label{fig:min-max-avg-bv-ex}
\end{figure}

We begin by calculating the brightness value (BV) for each image as the median brightness of the pixels in the face skin region.
We use the median rather than mean, given that some of the brightness value distributions exhibit significant skew. (This idea will be explored in-depth in Sec. \ref{experiment2}.) 
Example CF / AF images with the minimum, average, and maximum BV 
for that demographic
are shown in Fig. \ref{fig:min-max-avg-bv-ex}.

Then, for each mated pair, we calculate the brightness value difference (BVD) of its images $X, Y$ as:

\begin{equation}
\begin{split}
BV_X &= \text{Brightness Value of } X \\
BV_Y &= \text{Brightness Value of } Y \\
BVD_{X,Y} &= \lvert BV_X - BV_Y \rvert \\
\end{split}
\end{equation}

\begin{figure*}[!ht]
  \centering
  \includegraphics[width=.85\textwidth]{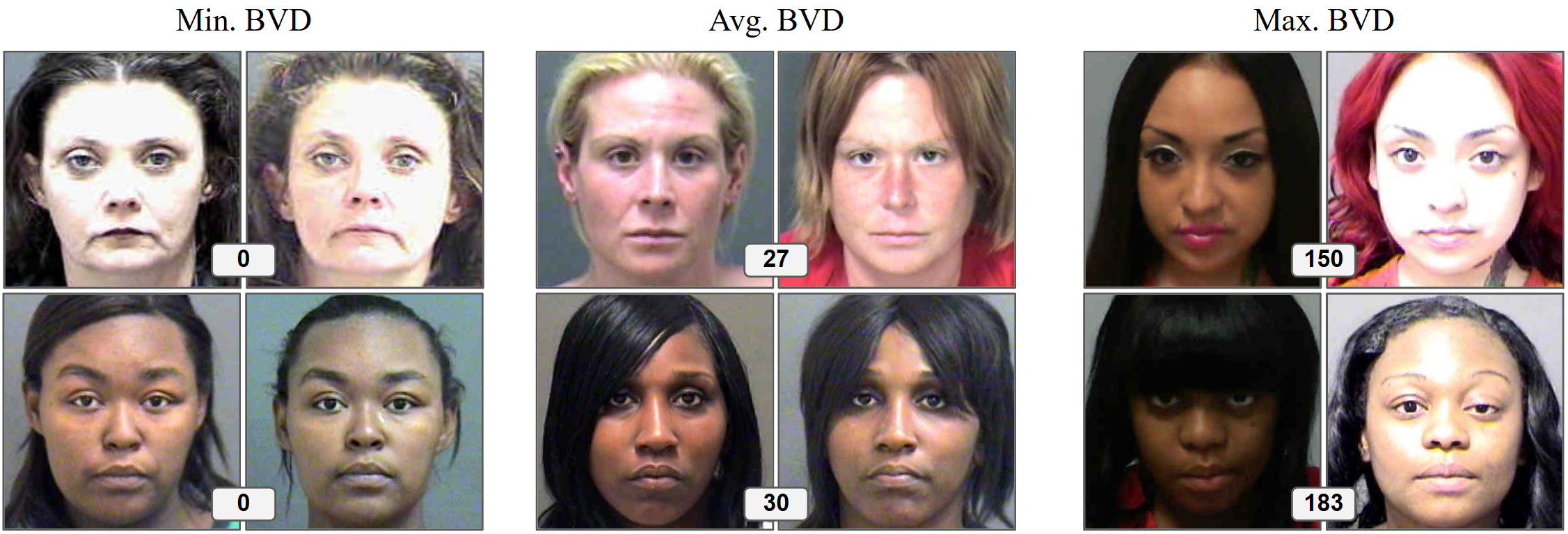}
  \caption{Example pairs with  the min, avg., and max BVDs.}
  \label{fig:min-max-avg-bv-diff}
\end{figure*}

Example CF / AF pairs with the minimum, average, and maximum BVD (calculated individually by demographic) are shown in Fig. \ref{fig:min-max-avg-bv-diff}.

\subsection{Balancing on BVD}
We begin by sorting all pairs from low to high BVD. For each CF pair, we attempt to find a unique AF pair with the same BVD. If there \textit{is} a same-BVD AF pair, we will ``match'' the two pairs. Note that this approach does not take into account the actual brightness values of a pair's individual images - just the difference between the two.

\begin{figure}[!ht]
  \centering
  \includegraphics[width=\columnwidth]{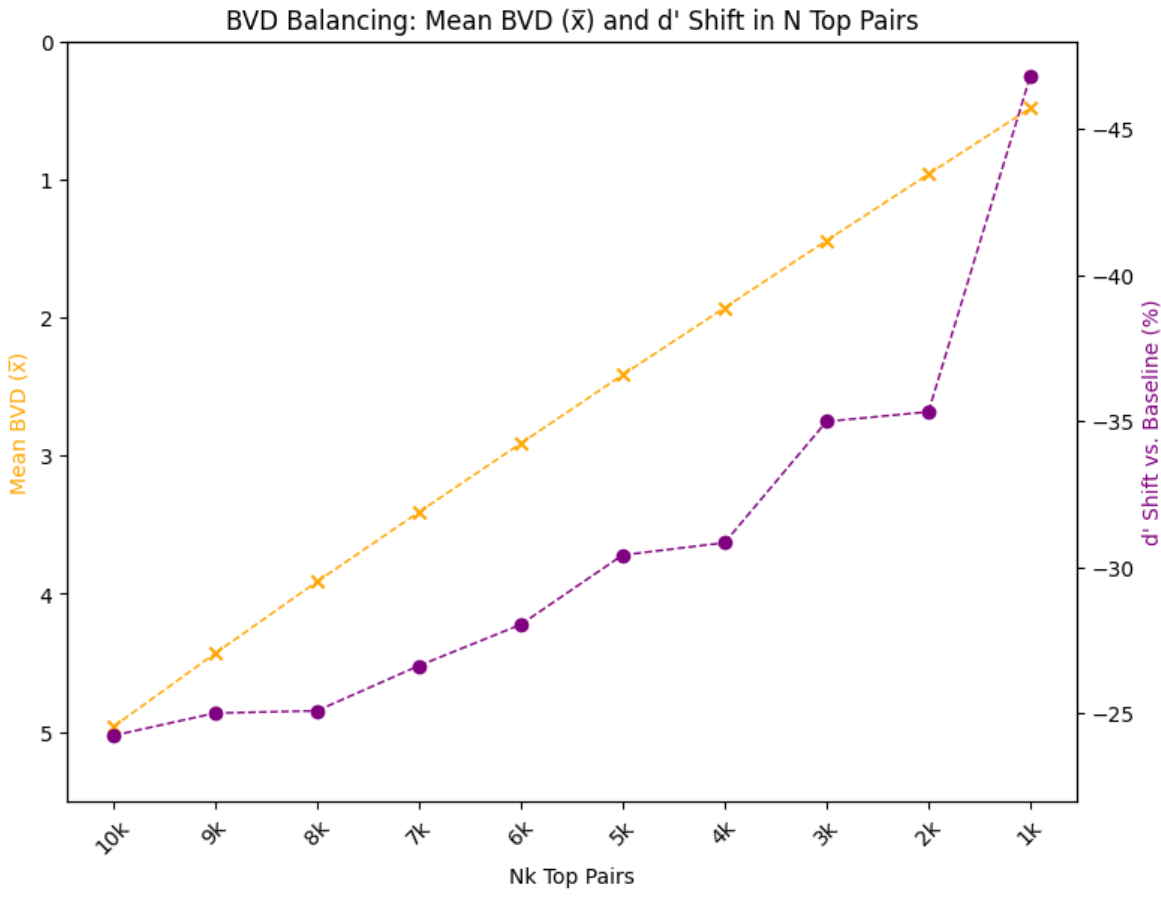}
  \caption{Relationship between mean BVD ($\bar{x}_b$) decrease and d' decrease.}
  \label{fig:bvd-bal-plot}
\end{figure}

Nearly all CF pairs can be ``matched'' with a unique AF pair. From the 33,735 total ``matched pairs'' ordered from low to high BVD, we can take the top N pairs. Tab. \ref{tab:bvd-res-short} and Fig. \ref{fig:bvd-bal-plot} highlight the results: as the BVD of the represented pairs approaches 0, the CF / AF  score  distributions become closer. 

\begin{table}[!ht]
\caption{N Top Pairs ordered by BVD.}
\label{tab:bvd-res-short}
\resizebox{\columnwidth}{!}{%
\begin{tabular}{|ccccc|}
\hline
\multicolumn{5}{|c|}{\cellcolor[HTML]{EFEFEF}\textbf{Bal. Factor: BVD}} \\ \hline
\multicolumn{1}{|c|}{\textbf{\begin{tabular}[c]{@{}c@{}}N Top \\ Pairs\end{tabular}}} & \multicolumn{1}{c|}{\textbf{Dem.}} & \multicolumn{1}{c|}{\textbf{\begin{tabular}[c]{@{}c@{}}Score $\bar{x}_{b}$\\ Shift\end{tabular}}} & \multicolumn{1}{c|}{\textbf{\begin{tabular}[c]{@{}c@{}}d'\\ Shift\end{tabular}}} & \textbf{\begin{tabular}[c]{@{}c@{}}BVD\\ $\bar{x}_{b}$\end{tabular}} \\ \hline
\multicolumn{1}{|c|}{} & \multicolumn{1}{c|}{CF} & \multicolumn{1}{c|}{3.7\%} & \multicolumn{1}{c|}{} &  \\ \cline{2-3}
\multicolumn{1}{|c|}{\multirow{-2}{*}{10k}} & \multicolumn{1}{c|}{AF} & \multicolumn{1}{c|}{2.0\%} & \multicolumn{1}{c|}{\multirow{-2}{*}{-24.2\%}} & \multirow{-2}{*}{5.0} \\ \hline
\multicolumn{1}{|c|}{} & \multicolumn{1}{c|}{CF} & \multicolumn{1}{c|}{4.0\%} & \multicolumn{1}{c|}{} &  \\ \cline{2-3}
\multicolumn{1}{|c|}{\multirow{-2}{*}{5k}} & \multicolumn{1}{c|}{AF} & \multicolumn{1}{c|}{2.1\%} & \multicolumn{1}{c|}{\multirow{-2}{*}{-30.4\%}} & \multirow{-2}{*}{2.4} \\ \hline
\multicolumn{1}{|c|}{} & \multicolumn{1}{c|}{CF} & \multicolumn{1}{c|}{4.2\%} & \multicolumn{1}{c|}{} &  \\ \cline{2-3}
\multicolumn{1}{|c|}{\multirow{-2}{*}{1k}} & \multicolumn{1}{c|}{AF} & \multicolumn{1}{c|}{1.4\%} & \multicolumn{1}{c|}{\multirow{-2}{*}{-46.8\%}} & \multirow{-2}{*}{0.5} \\ \hline
\end{tabular}%
}
\end{table}

\subsection{BVD Analysis}

Distributions of CF / AF pairs balanced such that for each CF pair, there is an AF pair with the same BVD, are shifted to higher scores and have a lower d' than the baseline distributions. As mean BVD consistently decreases, the d' shift vs. the baseline d' consistently (though not linearly) decreases. When taking the top 1k pairs (with a mean BVD of 0.5), the d' shift vs. the baseline d' is -46.8\%.

\section{Brightness Distribution Modality} \label{experiment2}

\subsection{Pre-Processing}
Each image's distribution of brightness pixels can be described as: unimodal (``Uni''), bimodal (``Bi''), or multimodal (``Multi''). (Our method for assigning modality is described in the Supplementary Material.)
Fig. \ref{fig:ex-uni-bi-images} shows an example of the same CF subject across three image instances: one Uni, one Bi, and one Multi. On each image's corresponding distribution, a red x marks the peak(s) determined according to our modality-assignment method. The y-axis indicates relative frequency of a given pixel value with respect to the rest of the face skin region. 

\begin{figure*}[!ht]
  \centering
  \includegraphics[width=.8\textwidth]{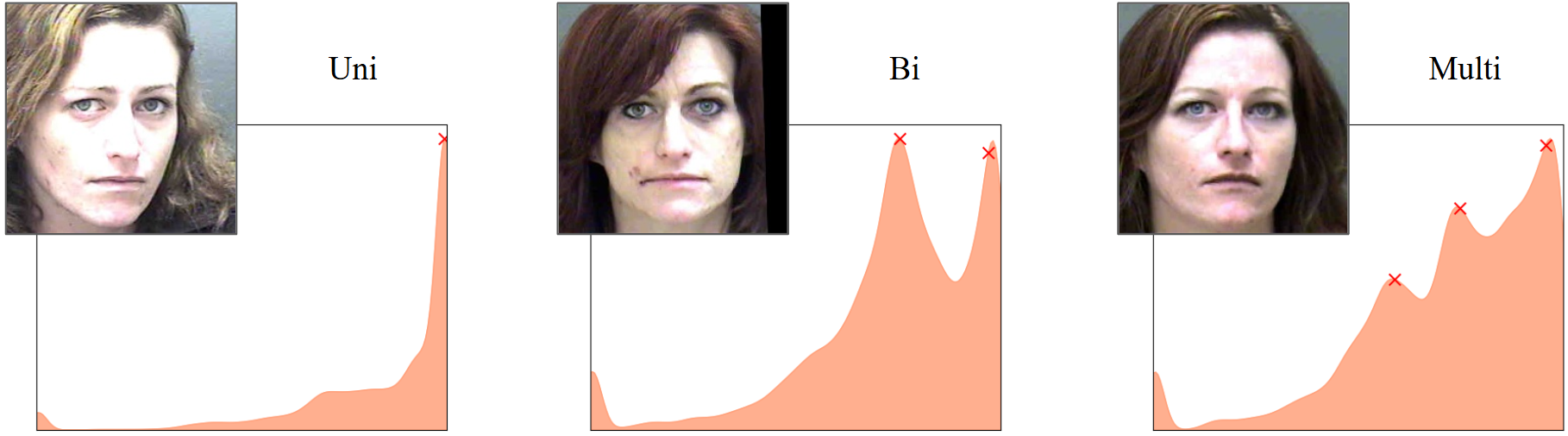}
  \caption{Example Uni / Bi / Multi images (and corresponding distributions) for the same CF subject.}
  \label{fig:ex-uni-bi-images}
\end{figure*}

After determining the brightness distribution modality (BDM) of each \textit{individual} image, we can label every mated pair according to the BDM of its two images. This labeling scheme allows for six possible pair types: UniUni (UU), BiBi (BB), MultiMulti (MM), UniBi (UB), UniMulti (UM), and BiMulti (BM). 

We can separate the CF / AF mated pairs into subsets according to each pair's type. Tab. \ref{tab:bdm-all} shows the baseline number of pairs of each type for the two demographics (``Num. Pairs''), along with the percentage of all pairs that each type makes up for the given demographic (``\% of All Pairs''). Note that the pairs of each type shown in Tab. \ref{tab:bdm-all} are not equal for the two demographics.

\begin{table}[!ht]
\caption{Details for all BDM pair types.}
\label{tab:bdm-all}
\resizebox{\columnwidth}{!}{%
\begin{tabular}{|c|c|c|c|c|c|}
\hline
\rowcolor[HTML]{EFEFEF} 
\textbf{\begin{tabular}[c]{@{}c@{}}Pair\\ Type\end{tabular}} & \textbf{Dem.} & \textbf{\begin{tabular}[c]{@{}c@{}}Num.\\ Pairs\end{tabular}} & \textbf{\begin{tabular}[c]{@{}c@{}}\% of \\ All Pairs\end{tabular}} & \textbf{\begin{tabular}[c]{@{}c@{}}Score $\bar{x}_{b}$\\ Shift\end{tabular}} & \textbf{\begin{tabular}[c]{@{}c@{}}d'\\ Shift\end{tabular}} \\ \hline
 & CF & 9949 & 29\% & -1.5\% &  \\ \cline{2-5}
\multirow{-2}{*}{UU} & AF & 14134 & 17\% & -0.6\% & \multirow{-2}{*}{15.6\%} \\ \hline
 & CF & 5247 & 16\% & 2.7\% &  \\ \cline{2-5}
\multirow{-2}{*}{BB} & AF & 17073 & 21\% & 1.1\% & \multirow{-2}{*}{-25.8\%} \\ \hline
 & CF & 590 & 2\% & 7.3\% &  \\ \cline{2-5}
\multirow{-2}{*}{MM} & AF & 3138 & 4\% & 2.4\% & \multirow{-2}{*}{-83.5\%} \\ \hline
 & CF & 11919 & 35\% & -1.3\% &  \\ \cline{2-5}
\multirow{-2}{*}{UB} & AF & 25334 & 31\% & -1.1\% & \multirow{-2}{*}{2.8\%} \\ \hline
 & CF & 2950 & 9\% & -0.4\% &  \\ \cline{2-5}
\multirow{-2}{*}{UM} & AF & 9523 & 12\% & -0.6\% & \multirow{-2}{*}{-0.5\%} \\ \hline
 & CF & 3085 & 9\% & 4.2\% &  \\ \cline{2-5}
\multirow{-2}{*}{BM} & AF & 13370 & 16\% & 1.1\% & \multirow{-2}{*}{-53.4\%} \\ \hline
\end{tabular}%
}
\end{table}

The distributions of some subsets (pairs that are UniUni and UniBi) have an \textit{increased} d' vs. the baseline d' (15.6\% and 2.5\%, respectively), while some subsets (pairs that are BiBi, MultiMulti, or BiMulti) have a \textit{decreased} d' (ranging from a 25.8-83.5\% decrease).

\subsection{Balancing on BDM}

With this in mind, we create balanced subsets consisting of different combinations of BiBi / MultiMulti / BiMulti pairs. We will consider three groupings: 
\begin{itemize}
    \item Pairs with only Bi and Multi images \\ \textit{BB, MM, \& BM Pairs}
    \item Pairs that include at least one Bi image \\ \textit{BB \& BM Pairs}
    \item Pairs that include at least one Multi image \\ \textit{MM \& BM Pairs}
\end{itemize}

For the first two groups, we can balance by taking 5k pairs each from the CF and AF distributions. For the third group, we are limited by the fact that MM pairs in general are a small proportion of each demographic's image set, and take only 3k pairs. 

When taking our subsets, since there is no inherent ordering, we shuffle the full set of pairs each time, and average across 10 trials to get the CF / AF mean scores and d' values. The results are given in Tab. \ref{tab:bdm-balancing}.

\begin{table}[!ht]
\caption{N Selected Pairs from each Non-Uni grouping.}
\label{tab:bdm-balancing}
\resizebox{\columnwidth}{!}{%
\begin{tabular}{|ccccc|}
\hline
\multicolumn{5}{|c|}{\cellcolor[HTML]{EFEFEF}\textbf{Bal. Factor: BDM}} \\ \hline
\multicolumn{1}{|c|}{\textbf{\begin{tabular}[c]{@{}c@{}}Pair\\ Types\end{tabular}}} & \multicolumn{1}{c|}{\textbf{\begin{tabular}[c]{@{}c@{}}N Sel.\\ Pairs\end{tabular}}} & \multicolumn{1}{c|}{\textbf{Dem.}} & \multicolumn{1}{c|}{\textbf{\begin{tabular}[c]{@{}c@{}}Score $\bar{x}_{b}$\\ Shift\end{tabular}}} & \textbf{\begin{tabular}[c]{@{}c@{}}d'\\ Shift\end{tabular}} \\ \hline
\multicolumn{1}{|c|}{} & \multicolumn{1}{c|}{} & \multicolumn{1}{c|}{CF} & \multicolumn{1}{c|}{3.5\%} &  \\ \cline{3-4}
\multicolumn{1}{|c|}{\multirow{-2}{*}{BB, MM, BM}} & \multicolumn{1}{c|}{\multirow{-2}{*}{5k}} & \multicolumn{1}{c|}{AF} & \multicolumn{1}{c|}{1.2\%} & \multirow{-2}{*}{-39.6\%} \\ \hline
\multicolumn{1}{|c|}{} & \multicolumn{1}{c|}{} & \multicolumn{1}{c|}{CF} & \multicolumn{1}{c|}{3.2\%} &  \\ \cline{3-4}
\multicolumn{1}{|c|}{\multirow{-2}{*}{BB, BM}} & \multicolumn{1}{c|}{\multirow{-2}{*}{5k}} & \multicolumn{1}{c|}{AF} & \multicolumn{1}{c|}{1.1\%} & \multirow{-2}{*}{-36.0\%} \\ \hline
\multicolumn{1}{|c|}{} & \multicolumn{1}{c|}{} & \multicolumn{1}{c|}{CF} & \multicolumn{1}{c|}{4.7\%} &  \\ \cline{3-4}
\multicolumn{1}{|c|}{\multirow{-2}{*}{MM, BM}} & \multicolumn{1}{c|}{\multirow{-2}{*}{3k}} & \multicolumn{1}{c|}{AF} & \multicolumn{1}{c|}{1.3\%} & \multirow{-2}{*}{-57.6\%} \\ \hline
\end{tabular}%
}
\end{table}

\subsection{BDM Analysis}
Balanced distributions excluding all pairs with a Uni image (i.e. only BB, MM, \& BM pairs) give a 39.6\% decrease in d'. Including only pairs with a Bi image (BB \& BM) gives a 36\% d' decrease. Including only pairs with a Multi image (MM \& BM) gives a 57.6\% d' decrease. In all three cases, the mean score of the balanced CF distribution improves almost 3x more than that of AF (3.2-4.7\% for CF vs. only 1.1-1.3\% for AF). 

Given the fact that pairs with at least one Uni image yield worse similarity scores for both groups, we might ask - what specifically about Uni images causes this? 
%
%
Tab. \ref{tab:med-bdm-pixvals} shows the median pixel value for each demographic and image type. Notice that the median values are least similar for CF / AF Uni images and most similar for Multi images, which perhaps helps explain why taking only UU pairs yields the highest d' increase (+15.6\%), and taking only MM pairs yields the highest decrease (-83.5\%) (as seen in Tab. \ref{tab:bdm-all}). 
\begin{table}[!ht]
\centering
\caption{Median pixel value for each BDM type.}
\label{tab:med-bdm-pixvals}
\begin{tabular}{c|c|c|c|}
\cline{2-4}
 & \textbf{Uni} & \textbf{Bi} & \textbf{Multi} \\ \hline
\multicolumn{1}{|c|}{\textbf{CF}} & 196 & 183 & 179 \\ \hline
\multicolumn{1}{|c|}{\textbf{AF}} & 118 & 132 & 137 \\ \hline
\end{tabular}
\end{table}

\begin{figure*}[!ht]
  \centering
  \includegraphics[width=.7\textwidth]{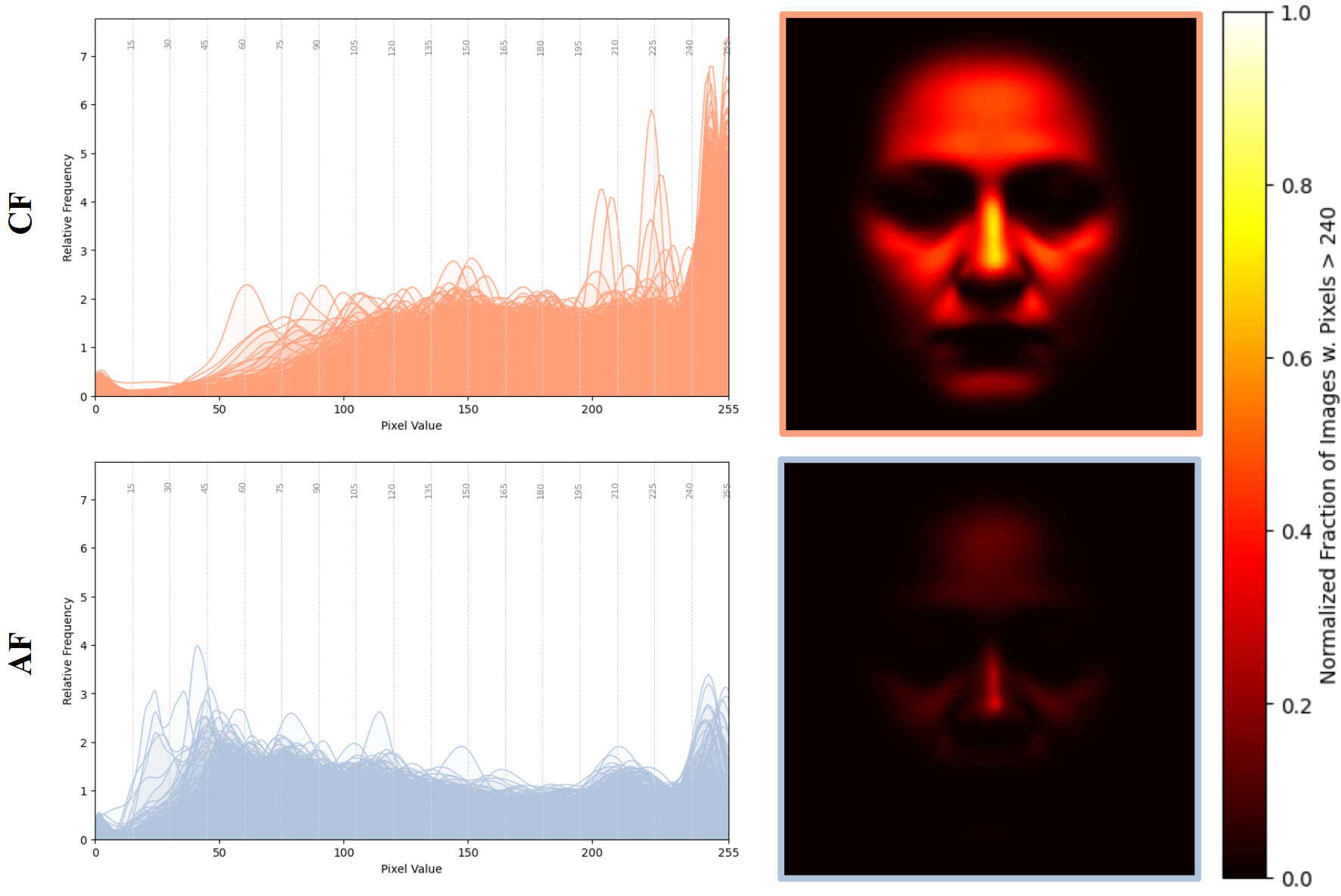}
  \caption{For Uni images, overlaid brightness distributions (left) and heatmaps highlighting the fraction of pixels with BVs greater than 240 (right). Each heatmap is normalized with respect to the number of Uni images for the given demographic.}
  \label{fig:heatmaps}
\end{figure*}

Fig. \ref{fig:heatmaps} shows histograms and heatmaps for CF / AF Uni Images.  The histograms show overlaid Uni distributions. Notice that for CF, the single peak causing the distributions to be classified as ``Uni'' seems to primarily occur for BVs 240-255 - these correspond to nearly-white pixels. The single peak for AF Uni images can also occur here (though at a sigfinicantly lower relative frequency), but can also occur at lower BVs, in the range of 30-90. Each heatmap overlays the pixels with BV $>$ 240 from all CF Uni / AF Uni images, respectively. 

\begin{figure}[!ht]
  \centering
  \includegraphics[width=\columnwidth]{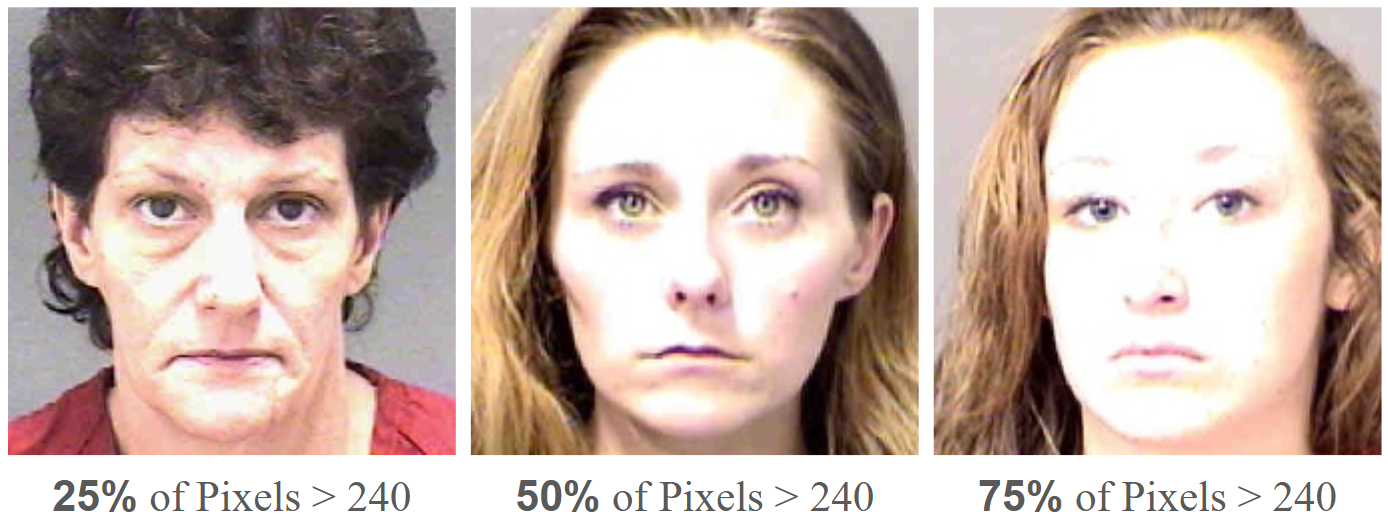}
  \caption{Ex. CF Uni images with \% of pixels $>$ 240.}
  \label{fig:cfuni-gt-240}
\end{figure}

Fig. \ref{fig:cfuni-gt-240} shows example CF images with increasing percentages of their face skin pixels $>$ 240. As this percentage increases, the face appears more overexposed - and facial feature information is lost / obscured. A total of 2197 CF Uni images (40\%) are at least as overexposed as the first example in Fig. \ref{fig:cfuni-gt-240}, and 411 images (7.5\%) are at least as overexposed as the second example.

\section{Brightness Distribution Similarity}\label{experiment3}

The previous experiment has given us the insight that balancing based on how the pixels representing brightness are distributed not only reduces the CF-AF accuracy gap, but also improves both underlying distributions (i.e. increases similarity scores). However, we also see that we cannot say a given modality means a ``universally worse'' image - there is a fundamental difference in CF and AF Uni images, for example. Since each demographic experiences better or worse scores for specific distribution shapes, we will now attempt to balance subsets by accounting for the similarity between both images in a CF pair and both images in an AF pair. We will compute similarity for these four images using intersection-over-union (IoU). IoU combines the idea of intersection with that of union, ``rewarding'' the areas in which they are similar (overlap) and ``penalizing'' the areas in which they differ (non-overlap). 

\subsection{Balancing on BD-IoU}

Here, we will consider each possible ``set'' of CF / AF Non-Uni pairs and compute the IoU of the brightness distributions (BD-IoU) across the two pairs. Algs. \ref{alg:bd-iou} and \ref{alg:iou-only} show this calculation. The goal of this experiment is to balance distributions such that for each CF pair, there is an AF pair for which the brightness distributions of the two CF images ($C_1,C_2$) have a high level of similarity (as represented by BD-IoU value) with those of the two AF images ($A_1,A_2$).

\RestyleAlgo{ruled}

\SetKwComment{Comment}{/* }{ */}

\begin{algorithm}[hbt!]
\caption{Compute BD-IoU of a Set}\label{alg:bd-iou}
\vspace{0.25em}
\KwData{$BD_{C_1}, BD_{C_2}, BD_{A_1}, BD_{A_2}$}
\KwResult{$BDIoU_{C_1,C_2,A_1,A_2}$}
\vspace{0.5em}
\tcp{Set 1}
$IoU_{C_1,A_1} \gets computeIoU(BD_{C_1}, BD_{A_1})$\;
$IoU_{C_2,A_2} \gets computeIoU(BD_{C_2}, BD_{A_2})$\;
$Avg_1 \gets mean(IoU_{C_1,A_1},IoU_{C_2,A_2})$\;
\vspace{0.5em}
\tcp{Set 2}
$IoU_{C_2,A_1} \gets computeIoU(BD_{C_2}, BD_{A_1})$\;
$IoU_{C_1,A_2} \gets computeIoU(BD_{C_1}, BD_{A_2})$\;
$Avg_2 \gets mean(IoU_{C_2,A_1},IoU_{C_1,A_2})$\;
\vspace{0.5em}
$BDIoU_{C_1,C_2,A_1,A_2} \gets max(Avg_1,Avg_2)$
\vspace{0.25em}
\end{algorithm}
\RestyleAlgo{ruled}

\SetKwComment{Comment}{/* }{ */}

\begin{algorithm}[hbt!]
\caption{Compute IoU of Two Images}\label{alg:iou-only}
\vspace{0.25em}
\KwData{$BD_{1}, BD_{2}$}
\KwResult{$IoU_{1,2}$}
\vspace{0.5em}
$intersection_{1,2} \gets sum(min(BD_{1},BD_{2}))$\;
$union_{1,2} \gets sum(max(BD_{1},BD_{2}))$\;
$IoU_{1,2} \gets intersection_{1,2}/union_{1,2}$\
\vspace{0.25em}
\end{algorithm}

\subsection{Pre-Processing}

After performing the calculation for all possible ``sets'' of CF / AF pairs, we can sort all sets from high to low BD-IoU. Finally, we construct our balanced subset by first adding the set that has the highest global BD-IoU value, and continuing to add sets of CF / AF pairs while ensuring uniqueness of both pairs. Ultimately, we can identify about 4,300 sets of unique CF / AF pairs, and from these we can now proceed as we did in the first experiment - taking the top N pairs ordered from high to low BD-IoU while incrementally decreasing N. The results of this experiment are shown in Tab. \ref{tab:no-uni-bdiou}.

\begin{table}[!ht]
\caption{N Top Pairs sorted from high to low BD-IoU.}
\label{tab:no-uni-bdiou}
\resizebox{\columnwidth}{!}{%
\begin{tabular}{|ccccc|}
\hline
\multicolumn{5}{|c|}{\textbf{Bal. Factor: BD-IoU}} \\ \hline
\multicolumn{1}{|c|}{\textbf{\begin{tabular}[c]{@{}c@{}}N Top\\ Pairs\end{tabular}}} & \multicolumn{1}{c|}{\textbf{Dem.}} & \multicolumn{1}{c|}{\textbf{\begin{tabular}[c]{@{}c@{}}Score $\bar{x}_{b}$\\ Shift\end{tabular}}} & \multicolumn{1}{c|}{\textbf{\begin{tabular}[c]{@{}c@{}}d'\\ Shift\end{tabular}}} & \textbf{\begin{tabular}[c]{@{}c@{}}BD-IoU\\$\bar{x}_{b}$\end{tabular}} \\ \hline
\multicolumn{1}{|c|}{\multirow{2}{*}{4k}} & \multicolumn{1}{c|}{CF} & \multicolumn{1}{c|}{3.64\%} & \multicolumn{1}{c|}{\multirow{2}{*}{-32.68\%}} & \multirow{2}{*}{0.785} \\ \cline{2-3}
\multicolumn{1}{|c|}{} & \multicolumn{1}{c|}{AF} & \multicolumn{1}{c|}{1.56\%} & \multicolumn{1}{c|}{} &  \\ \hline
\multicolumn{1}{|c|}{\multirow{2}{*}{3k}} & \multicolumn{1}{c|}{CF} & \multicolumn{1}{c|}{4.16\%} & \multicolumn{1}{c|}{\multirow{2}{*}{-31.51\%}} & \multirow{2}{*}{0.805} \\ \cline{2-3}
\multicolumn{1}{|c|}{} & \multicolumn{1}{c|}{AF} & \multicolumn{1}{c|}{2.06\%} & \multicolumn{1}{c|}{} &  \\ \hline
\multicolumn{1}{|c|}{\multirow{2}{*}{2k}} & \multicolumn{1}{c|}{CF} & \multicolumn{1}{c|}{4.57\%} & \multicolumn{1}{c|}{\multirow{2}{*}{-26.78\%}} & \multirow{2}{*}{0.821} \\ \cline{2-3}
\multicolumn{1}{|c|}{} & \multicolumn{1}{c|}{AF} & \multicolumn{1}{c|}{2.63\%} & \multicolumn{1}{c|}{} &  \\ \hline
\multicolumn{1}{|c|}{\multirow{2}{*}{1k}} & \multicolumn{1}{c|}{CF} & \multicolumn{1}{c|}{5.94\%} & \multicolumn{1}{c|}{\multirow{2}{*}{-30.60\%}} & \multirow{2}{*}{0.836} \\ \cline{2-3}
\multicolumn{1}{|c|}{} & \multicolumn{1}{c|}{AF} & \multicolumn{1}{c|}{3.67\%} & \multicolumn{1}{c|}{} &  \\ \hline
\end{tabular}%
}
\end{table}

\subsection{BD-IoU Analysis}
Distributions of CF / AF pairs balanced such that for each CF pair, its images' distributions have at least 78.5\% similarity to the images of some AF pair, are shifted to higher scores and have a lower d' than the baseline distributions. As mean BD-IoU consistently increases, the d' shift vs. the baseline d' decreases as well - but this decrease is \textit{not} consistent. For the top 1k pairs (with BD-IoU $\bar{x}_b$ = 0.836), the d' shift vs. the baseline is -30.6\%. 
\section{Discussion}\label{discussion}

In this work, we explore how the image quality factor of illumination (as related to brightness of the face region) impacts mated score distributions for Caucasian and African American females. We balance CF and AF distributions by requiring that, for a given mated pair, its two images are similarly illuminated (Exp. 1) or well illuminated (Exp. 2). These requirements are combined in Exp. 3, in which we ensure that for a given well illuminated CF pair, there exists a ``counterpart'' well illuminated AF pair whose two images are similarly illuminated to those of the CF pair.

\subsection{Brightness Value Difference}

In the first experiment, we measure the brightness value (BV) of a given image as the median pixel value of the extracted face skin region, then compute the brightness value difference (BVD) of the images in each pair. We balanced subsets of CF / AF pairs by (1) sorting CF pairs from low to high BVD, (2) for each CF pair, finding a unique AF pair with the same BVD, (3) taking the top N pairs (ordered from low to high BVD). 

Interestingly, this method does not take into account the brightness values of individual images, and only requires that the brightness \textit{difference} between any two images comprising a pair is minimized (i.e. that they are similarly illuminated). When we require that the brightness difference is $\leq$ 0.5 (as it was when taking the 1k Top Pairs), we can decrease the CF-AF accuracy gap by at least 46.80\%.  

\subsection{Brightness Distribution Modality} 

In the second two experiments, we interpret the brightness of a given image in terms of the distribution of all pixel values from the extracted face skin region. The ``brightness distribution'' (BD) of each image can be labeled with respect to its modality (unimodal, bimodal, or multimodal), and each mated pair is labeled according to the modalities of its two images. We find that unimodality tends to indicate poorly illuminated images, while bi-/multimodality indicates well illuminated images. Thus, we balance distributions by taking the same number of CF / AF pairs that only include Bi / Multi images.

\subsection{Brightness Distribution IoU}

In order to determine how similarly illuminated two mated pairs are, we can calculate the intersection-over-union of their brightness distributions (BD-IoU).
BD-IoU provides a discrete-value measure of similarity between the brightness distributions of the four images across two mated pairs. In the corresponding balanced distributions, any pairs with Uni images (i.e. poorly illuminated images) are excluded. Then, for each represented CF pair, the brightness distributions of its two images have a high level of similarity with those of the two images in a unique AF pair.

Table \ref{tab:comparing-all} compares the ``best'' results from each experiment. Note that for the BVD and BDIOU experiment, where there is an ordering (e.g. low to high BVD), we show the score ($\bar{x}_b$) shift / d' shift for the top 1k pairs. For the BDM experiment, we simply select 3k CF pairs and 3k AF pairs, where all pairs are either MultiMulti or BiMulti.

\begin{table}[!ht]
\caption{Comparing the results of each experiment w.r.t. CF / AF mean score ($\bar{x}_b$) shift and d' shift.}
\label{tab:comparing-all}
\resizebox{\columnwidth}{!}{%
\begin{tabular}{|c|c|c|c|c|}
\hline
\textbf{\begin{tabular}[c]{@{}c@{}}Bal. \\ Factor\end{tabular}} & \textbf{N Pairs} & \textbf{Dem} & \textbf{\begin{tabular}[c]{@{}c@{}}Score $\bar{x}_{b}$\\ Shift\end{tabular}} & \textbf{\begin{tabular}[c]{@{}c@{}}d'\\ Shift\end{tabular}} \\ \hline
\multirow{2}{*}{BVD} & \multirow{2}{*}{1k Top} & CF & 4.19\% & \multirow{2}{*}{-46.80\%} \\ \cline{3-4}
 &  & AF & 1.42\% &  \\ \hline
\multirow{2}{*}{BDM} & \multirow{2}{*}{\begin{tabular}[c]{@{}c@{}}3k Sel.\\ \textit{MM, BM}\end{tabular}} & CF & 4.69\% & \multirow{2}{*}{-57.59\%} \\ \cline{3-4}
 &  & AF & 1.30\% &  \\ \hline
\multirow{2}{*}{BDIOU} & \multirow{2}{*}{1k Top} & CF & 5.94\% & \multirow{2}{*}{-30.60\%} \\ \cline{3-4}
 &  & AF & 3.67\% &  \\ \hline
\end{tabular}%
}
\end{table}

For all three experiments, we accomplished our goal of decreasing the d' (i.e. reducing the gap between the balanced CF-AF distributions) while also increasing the mean similarity score of each individual distribution. Note that there is a tradeoff: if both balanced distributions shift significantly to higher similarity scores (as in the BD-IoU experiment), then the d' may not decrease as much as cases where the CF distribution's shift is much more significant than that of AF (as in BVD and BDM). 
\section{Conclusion}\label{conclusion}

In order to understand the demographic differentials in face recognition accuracy, it is essential to understand the role of image quality. In this work, we explore the image quality factor of illumination - specifically, the brightness of the face skin region. 

All three experiments we performed yielded positive results, both in terms of decreasing the performance gap between CF-AF distributions and in improving the scores of the individual distributions (thereby decreasing False Non-Match Rate). 

We find that we can decrease the performance gap $\sim$30-58\% by ensuring that one of the following conditions is met: \textbf{(1)} the difference in brightness values of two images in a pair is less than 1, \textbf{(2)} no pairs include images with unimodal distributions of brightness values (which appears to indicate overexposure for CF images), or \textbf{(3)} there is around 80\% similarity in the brightness distributions of two images in a given CF pair with the brightness distributions of two images in at least one AF pair. 

Ultimately, these findings point to the fact that it is essential to ensure at acquisition time that images are both similarly illuminated and well illuminated across demographics - for images of all subjects individually, and for images of the same person comprising mated pairs. 
This should be possible for operational scenarios where images are taken under controlled conditions, such as driver's license photos, passport photos, mugshots and the like.
The fundamental advantages that result are higher average accuracy within each demographic, and lower accuracy difference across demographics.

{\small
\bibliographystyle{ieee_fullname}
\bibliography{egbib}
}

\clearpage
\onecolumn
\begin{center}
    \Large \textbf{Supplementary Material}
\end{center}
\section*{Brightness Value Difference (BVD)}

\begin{figure*}[!ht]
    \centering
    \begin{subfigure}[b]{0.455\textwidth}
        \centering
        \includegraphics[width=\textwidth]{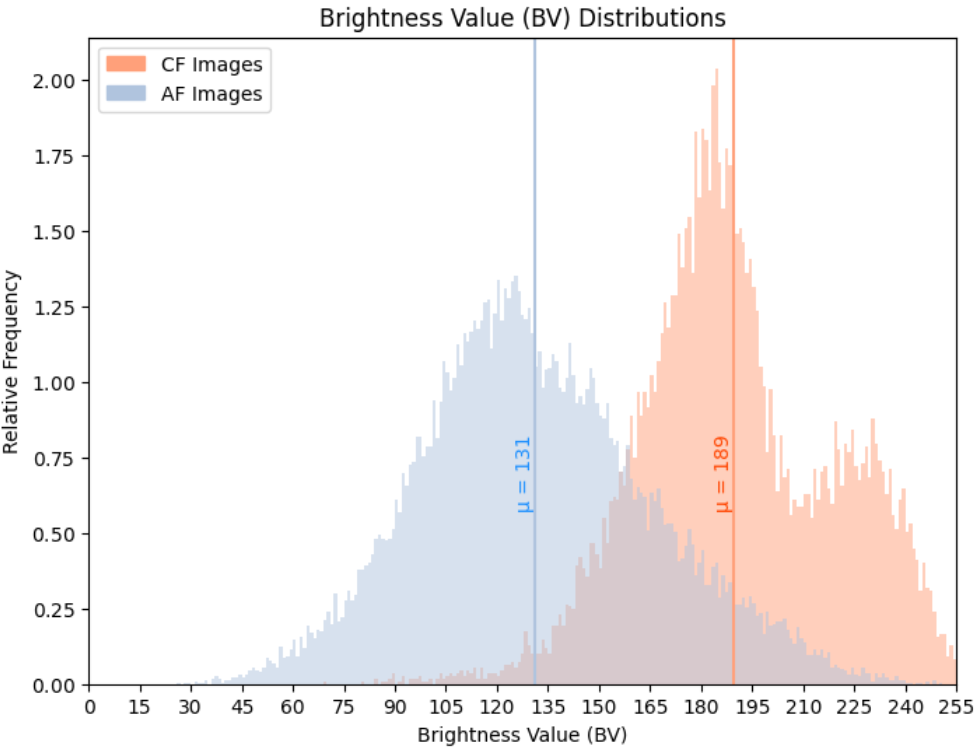} 
        \caption{BVs for CF / AF Images.}
        \label{fig:bv-dists}
    \end{subfigure}
    \hfill
    \begin{subfigure}[b]{0.45\textwidth}
        \centering
        \includegraphics[width=\textwidth]{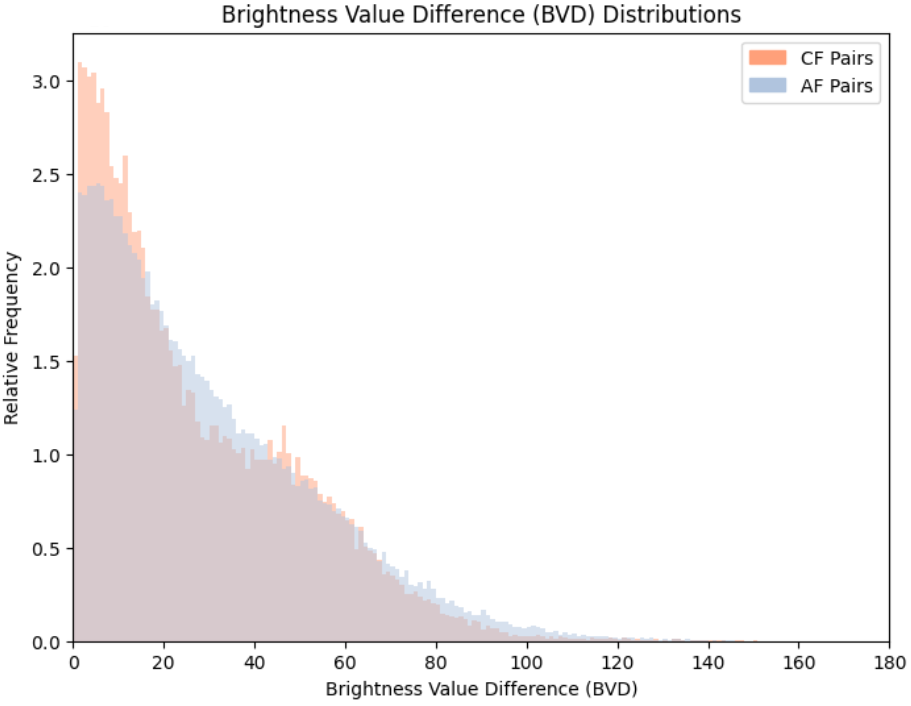} 
        \caption{BVDs for CF / AF Pairs.}
        \label{fig:bvd-dists}
    \end{subfigure}
    \caption{BV and BVD distributions.}
    \label{fig:bv-bvd-dists}
\end{figure*}

Each image is assigned a brightness value (BV) based on the median pixel value of the face skin region. Fig. \ref{fig:bv-dists} shows the distributions of BV values for CF / AF images. We use the BV values to calculate BVD for each pair. The distributions of BVD values for CF / AF pairs are shown in Fig. \ref{fig:bvd-dists}.

The full results of the BVD experiment are shown in Tab. \ref{tab:bvd-full-res}. For each balanced N Top Pairs subset, we report the mean similarity score (``Score $\bar{x}_b$''), shift in $\bar{x}_b$ vs. the baseline mean score (``Score $\bar{x}_b$'' Shift), d' between CF-AF, d' shift vs. the baseline d', as well as mean / std. dev / min. / max. BVD value for each subset.
\begin{table}[!ht]
\centering
\caption{Full results for BVD Experiment.}
\label{tab:bvd-full-res}
\begin{tabular}{|cccccccccc|}
\hline
\multicolumn{10}{|c|}{\textbf{Balancing Factor: Brightness Value Difference (BVD)}} \\ \hline
\multicolumn{1}{|c|}{\textbf{\begin{tabular}[c]{@{}c@{}}N Top \\ Pairs\end{tabular}}} & \multicolumn{1}{c|}{\textbf{Dem}} & \multicolumn{1}{c|}{\textbf{\begin{tabular}[c]{@{}c@{}}Score \\ $\bar{x}_{b}$\end{tabular}}} & \multicolumn{1}{c|}{\textbf{\begin{tabular}[c]{@{}c@{}}Score $\bar{x}_{b}$\\ Shift\end{tabular}}} & \multicolumn{1}{c|}{\textbf{d'}} & \multicolumn{1}{c|}{\textbf{\begin{tabular}[c]{@{}c@{}}d'\\ Shift\end{tabular}}} & \multicolumn{1}{c|}{\textbf{\begin{tabular}[c]{@{}c@{}}BVD \\ $\bar{x}_{b}$\end{tabular}}} & \multicolumn{1}{c|}{\textbf{\begin{tabular}[c]{@{}c@{}}BVD \\ STD\end{tabular}}} & \multicolumn{1}{c|}{\textbf{\begin{tabular}[c]{@{}c@{}}BVD\\ Min\end{tabular}}} & \textbf{\begin{tabular}[c]{@{}c@{}}BVD\\ Max\end{tabular}} \\ \hline
\multicolumn{1}{|c|}{\multirow{2}{*}{10k}} & \multicolumn{1}{c|}{CF} & \multicolumn{1}{c|}{0.7468} & \multicolumn{1}{c|}{3.66\%} & \multicolumn{1}{c|}{\multirow{2}{*}{0.3925}} & \multicolumn{1}{c|}{\multirow{2}{*}{-24.23\%}} & \multicolumn{1}{c|}{\multirow{2}{*}{4.965}} & \multicolumn{1}{c|}{\multirow{2}{*}{2.977}} & \multicolumn{1}{c|}{\multirow{2}{*}{0}} & \multirow{2}{*}{10} \\ \cline{2-4}
\multicolumn{1}{|c|}{} & \multicolumn{1}{c|}{AF} & \multicolumn{1}{c|}{0.7758} & \multicolumn{1}{c|}{2.03\%} & \multicolumn{1}{c|}{} & \multicolumn{1}{c|}{} & \multicolumn{1}{c|}{} & \multicolumn{1}{c|}{} & \multicolumn{1}{c|}{} &  \\ \hline
\multicolumn{1}{|c|}{\multirow{2}{*}{9k}} & \multicolumn{1}{c|}{CF} & \multicolumn{1}{c|}{0.7473} & \multicolumn{1}{c|}{3.73\%} & \multicolumn{1}{c|}{\multirow{2}{*}{0.3885}} & \multicolumn{1}{c|}{\multirow{2}{*}{-25.00\%}} & \multicolumn{1}{c|}{\multirow{2}{*}{4.434}} & \multicolumn{1}{c|}{\multirow{2}{*}{2.645}} & \multicolumn{1}{c|}{\multirow{2}{*}{0}} & \multirow{2}{*}{9} \\ \cline{2-4}
\multicolumn{1}{|c|}{} & \multicolumn{1}{c|}{AF} & \multicolumn{1}{c|}{0.776} & \multicolumn{1}{c|}{2.05\%} & \multicolumn{1}{c|}{} & \multicolumn{1}{c|}{} & \multicolumn{1}{c|}{} & \multicolumn{1}{c|}{} & \multicolumn{1}{c|}{} &  \\ \hline
\multicolumn{1}{|c|}{\multirow{2}{*}{8k}} & \multicolumn{1}{c|}{CF} & \multicolumn{1}{c|}{0.7474} & \multicolumn{1}{c|}{3.75\%} & \multicolumn{1}{c|}{\multirow{2}{*}{0.3881}} & \multicolumn{1}{c|}{\multirow{2}{*}{-25.08\%}} & \multicolumn{1}{c|}{\multirow{2}{*}{3.914}} & \multicolumn{1}{c|}{\multirow{2}{*}{2.327}} & \multicolumn{1}{c|}{\multirow{2}{*}{0}} & \multirow{2}{*}{8} \\ \cline{2-4}
\multicolumn{1}{|c|}{} & \multicolumn{1}{c|}{AF} & \multicolumn{1}{c|}{0.7761} & \multicolumn{1}{c|}{2.06\%} & \multicolumn{1}{c|}{} & \multicolumn{1}{c|}{} & \multicolumn{1}{c|}{} & \multicolumn{1}{c|}{} & \multicolumn{1}{c|}{} &  \\ \hline
\multicolumn{1}{|c|}{\multirow{2}{*}{7k}} & \multicolumn{1}{c|}{CF} & \multicolumn{1}{c|}{0.7482} & \multicolumn{1}{c|}{3.86\%} & \multicolumn{1}{c|}{\multirow{2}{*}{0.3801}} & \multicolumn{1}{c|}{\multirow{2}{*}{-26.62\%}} & \multicolumn{1}{c|}{\multirow{2}{*}{3.411}} & \multicolumn{1}{c|}{\multirow{2}{*}{2.031}} & \multicolumn{1}{c|}{\multirow{2}{*}{0}} & \multirow{2}{*}{7} \\ \cline{2-4}
\multicolumn{1}{|c|}{} & \multicolumn{1}{c|}{AF} & \multicolumn{1}{c|}{0.7763} & \multicolumn{1}{c|}{2.09\%} & \multicolumn{1}{c|}{} & \multicolumn{1}{c|}{} & \multicolumn{1}{c|}{} & \multicolumn{1}{c|}{} & \multicolumn{1}{c|}{} &  \\ \hline
\multicolumn{1}{|c|}{\multirow{2}{*}{6k}} & \multicolumn{1}{c|}{CF} & \multicolumn{1}{c|}{0.7484} & \multicolumn{1}{c|}{3.89\%} & \multicolumn{1}{c|}{\multirow{2}{*}{0.3728}} & \multicolumn{1}{c|}{\multirow{2}{*}{-28.03\%}} & \multicolumn{1}{c|}{\multirow{2}{*}{2.913}} & \multicolumn{1}{c|}{\multirow{2}{*}{1.743}} & \multicolumn{1}{c|}{\multirow{2}{*}{0}} & \multirow{2}{*}{6} \\ \cline{2-4}
\multicolumn{1}{|c|}{} & \multicolumn{1}{c|}{AF} & \multicolumn{1}{c|}{0.776} & \multicolumn{1}{c|}{2.05\%} & \multicolumn{1}{c|}{} & \multicolumn{1}{c|}{} & \multicolumn{1}{c|}{} & \multicolumn{1}{c|}{} & \multicolumn{1}{c|}{} &  \\ \hline
\multicolumn{1}{|c|}{\multirow{2}{*}{5k}} & \multicolumn{1}{c|}{CF} & \multicolumn{1}{c|}{0.7494} & \multicolumn{1}{c|}{4.03\%} & \multicolumn{1}{c|}{\multirow{2}{*}{0.3605}} & \multicolumn{1}{c|}{\multirow{2}{*}{-30.41\%}} & \multicolumn{1}{c|}{\multirow{2}{*}{2.417}} & \multicolumn{1}{c|}{\multirow{2}{*}{1.457}} & \multicolumn{1}{c|}{\multirow{2}{*}{0}} & \multirow{2}{*}{5} \\ \cline{2-4}
\multicolumn{1}{|c|}{} & \multicolumn{1}{c|}{AF} & \multicolumn{1}{c|}{0.7761} & \multicolumn{1}{c|}{2.06\%} & \multicolumn{1}{c|}{} & \multicolumn{1}{c|}{} & \multicolumn{1}{c|}{} & \multicolumn{1}{c|}{} & \multicolumn{1}{c|}{} &  \\ \hline
\multicolumn{1}{|c|}{\multirow{2}{*}{4k}} & \multicolumn{1}{c|}{CF} & \multicolumn{1}{c|}{0.7495} & \multicolumn{1}{c|}{4.04\%} & \multicolumn{1}{c|}{\multirow{2}{*}{0.3583}} & \multicolumn{1}{c|}{\multirow{2}{*}{-30.83\%}} & \multicolumn{1}{c|}{\multirow{2}{*}{1.931}} & \multicolumn{1}{c|}{\multirow{2}{*}{1.189}} & \multicolumn{1}{c|}{\multirow{2}{*}{0}} & \multirow{2}{*}{4} \\ \cline{2-4}
\multicolumn{1}{|c|}{} & \multicolumn{1}{c|}{AF} & \multicolumn{1}{c|}{0.7761} & \multicolumn{1}{c|}{2.06\%} & \multicolumn{1}{c|}{} & \multicolumn{1}{c|}{} & \multicolumn{1}{c|}{} & \multicolumn{1}{c|}{} & \multicolumn{1}{c|}{} &  \\ \hline
\multicolumn{1}{|c|}{\multirow{2}{*}{3k}} & \multicolumn{1}{c|}{CF} & \multicolumn{1}{c|}{0.7497} & \multicolumn{1}{c|}{4.07\%} & \multicolumn{1}{c|}{\multirow{2}{*}{0.3368}} & \multicolumn{1}{c|}{\multirow{2}{*}{-34.98\%}} & \multicolumn{1}{c|}{\multirow{2}{*}{1.445}} & \multicolumn{1}{c|}{\multirow{2}{*}{0.928}} & \multicolumn{1}{c|}{\multirow{2}{*}{0}} & \multirow{2}{*}{3} \\ \cline{2-4}
\multicolumn{1}{|c|}{} & \multicolumn{1}{c|}{AF} & \multicolumn{1}{c|}{0.7751} & \multicolumn{1}{c|}{1.93\%} & \multicolumn{1}{c|}{} & \multicolumn{1}{c|}{} & \multicolumn{1}{c|}{} & \multicolumn{1}{c|}{} & \multicolumn{1}{c|}{} &  \\ \hline
\multicolumn{1}{|c|}{\multirow{2}{*}{2k}} & \multicolumn{1}{c|}{CF} & \multicolumn{1}{c|}{0.749} & \multicolumn{1}{c|}{3.97\%} & \multicolumn{1}{c|}{\multirow{2}{*}{0.3351}} & \multicolumn{1}{c|}{\multirow{2}{*}{-35.31\%}} & \multicolumn{1}{c|}{\multirow{2}{*}{0.964}} & \multicolumn{1}{c|}{\multirow{2}{*}{0.69}} & \multicolumn{1}{c|}{\multirow{2}{*}{0}} & \multirow{2}{*}{2} \\ \cline{2-4}
\multicolumn{1}{|c|}{} & \multicolumn{1}{c|}{AF} & \multicolumn{1}{c|}{0.7742} & \multicolumn{1}{c|}{1.81\%} & \multicolumn{1}{c|}{} & \multicolumn{1}{c|}{} & \multicolumn{1}{c|}{} & \multicolumn{1}{c|}{} & \multicolumn{1}{c|}{} &  \\ \hline
\multicolumn{1}{|c|}{\multirow{2}{*}{1k}} & \multicolumn{1}{c|}{CF} & \multicolumn{1}{c|}{0.7506} & \multicolumn{1}{c|}{4.19\%} & \multicolumn{1}{c|}{\multirow{2}{*}{0.2756}} & \multicolumn{1}{c|}{\multirow{2}{*}{-46.80\%}} & \multicolumn{1}{c|}{\multirow{2}{*}{0.486}} & \multicolumn{1}{c|}{\multirow{2}{*}{0.5}} & \multicolumn{1}{c|}{\multirow{2}{*}{0}} & \multirow{2}{*}{1} \\ \cline{2-4}
\multicolumn{1}{|c|}{} & \multicolumn{1}{c|}{AF} & \multicolumn{1}{c|}{0.7712} & \multicolumn{1}{c|}{1.42\%} & \multicolumn{1}{c|}{} & \multicolumn{1}{c|}{} & \multicolumn{1}{c|}{} & \multicolumn{1}{c|}{} & \multicolumn{1}{c|}{} &  \\ \hline
\end{tabular}
\end{table}
\clearpage

\section*{Brightness Distribution Modality (BDM)}

In order to provide modality labels to each image, we analyzed its distribution using parameters for smoothing window (SW) and relative threshold (RT). 

First, we overlaid a smooth curve on top of the binned histogram of pixel values (representing the face region). The SW parameter determines the number of adjacent bins to average over when smoothing. A smaller SW value (e.g. 1-2) is more reflective of the original histogram shape, allowing for detection of more subtle features. A larger SW value (e.g. 9-10) yields a smoother curve that may reduce noise, but may also obscure smaller peaks in the data.

The RT parameter determines the minimum height a peak occurring in the smooth curve must be to be considered ``significant''. Its value is given as a fraction of the maximum count in the smoothed histogram. A lower RT value (e.g. 0.05) is useful for data with many minor (but important) fluctuations, but may detect noise as peaks. A higher RT value (e.g. 0.5) focuses on the distribution's major features. We use the RT value to label peaks on the smoothed curve. 

We tested multiple combinations of SW / RT value and manually checked the results. We ultimately selected SW = 4 and RT = 0.5.

\begin{figure*}[!ht]
  \centering
  \includegraphics[width=\textwidth]{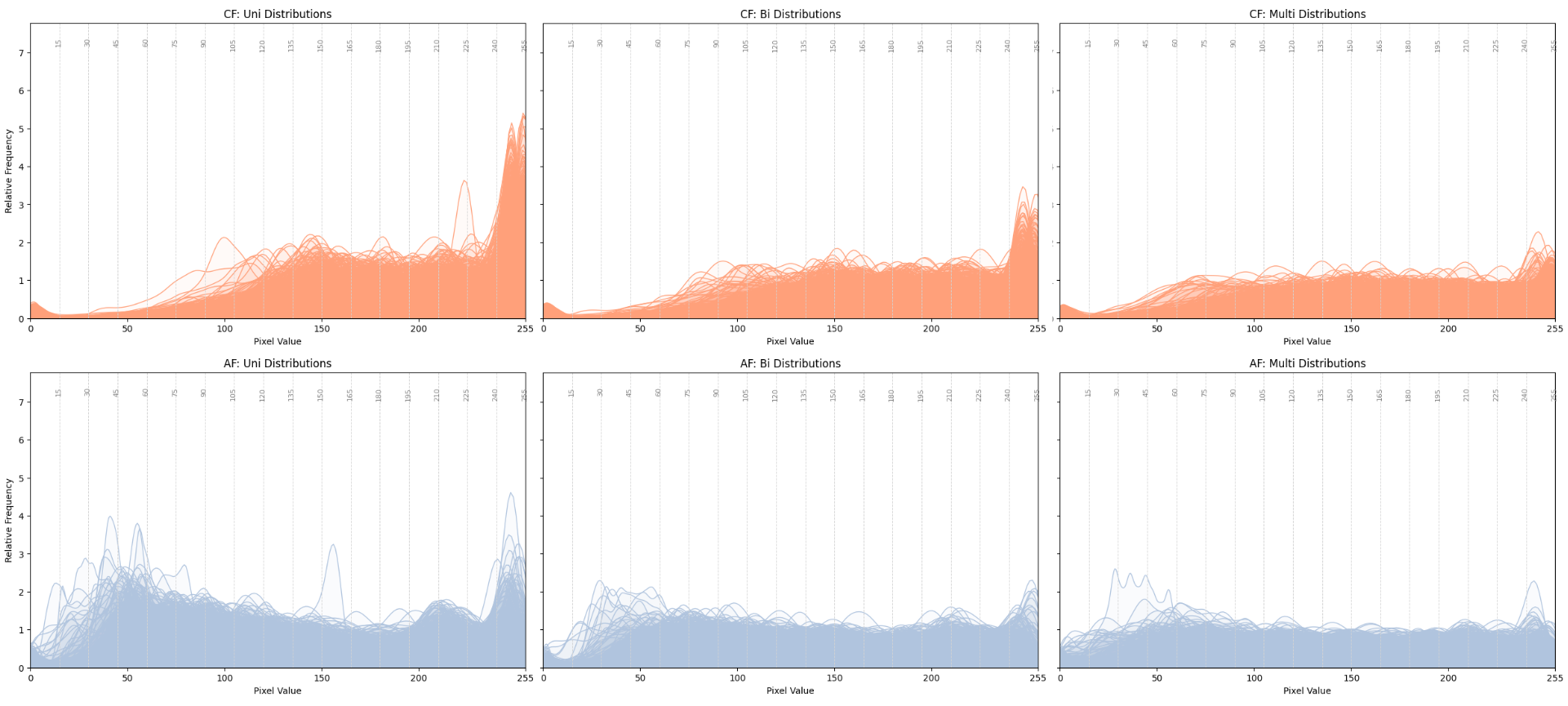}
  \caption{Relative frequency distributions of pixel values for Uni, Bi, and Multi images.}
  \label{fig:unibimulti}
\end{figure*}

The plots in Fig. \ref{fig:unibimulti} overlay the pixel distributions of Uni / Bi / Multi images. All plots have the same scaling for the y-axis (relative frequency).

\section*{Comparison of Best Results}
The plots on the next page show the baseline and balanced distributions corresponding to Table \ref{tab:comparing-all}.
\begin{figure*}[!ht]
  \centering
  \includegraphics[width=.7\textwidth]{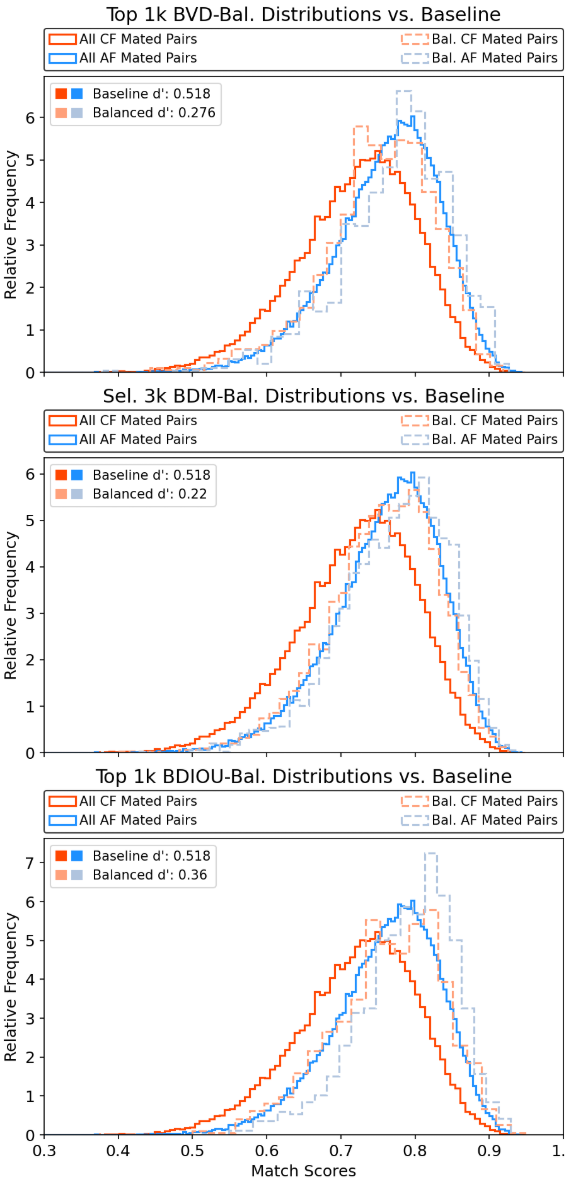}
\end{figure*}
\end{document}